\documentclass[letterpaper, 10 pt, conference]{ieeeconf}  

\IEEEoverridecommandlockouts                              
\usepackage[utf8]{inputenc}
\usepackage{color}
\usepackage{amsmath}
\usepackage{amsfonts}
\usepackage{amssymb}
\usepackage{booktabs}
\usepackage{paralist}
\usepackage{graphicx}
\usepackage{multirow}
\usepackage{subcaption}

\graphicspath{ {figures/} }
\usepackage[ruled]{algorithm2e}
\usepackage{algpseudocode} 

\DeclareMathOperator*{\argmin}{arg\,min}

\newlength\mylen
\newcommand\myinput[1]{%
  \settowidth\mylen{\KwIn{}}%
  \setlength\hangindent{\mylen}%
  \hspace*{\mylen}#1\\}
 
\title{\Large \bf Stochastic Functional Gradient Path Planning in Occupancy Maps}

\author{Gilad Francis, Lionel Ott and Fabio Ramos{$^*$}
\thanks{$^*$ Gilad Francis, Lionel Ott and Fabio Ramos are with The School of Information Technologies, University of Sydney, Australia
        {\tt\small gfra8070@uni.sydney.edu.au}}%
}
  
\begin{document}

\maketitle
\thispagestyle{empty}
\pagestyle{empty}

\begin{abstract}

Planning safe paths is a major building block in robot autonomy. It has been an active field of research for several decades, with a plethora of planning methods. Planners can be generally categorised as either trajectory optimisers or sampling-based planners. The latter is the predominat planning paradigm for occupancy maps. Trajectory optimisation entails major algorithmic changes to tackle contextual information gaps caused by incomplete sensor coverage of the map. However, the benefits are substantial, as trajectory optimisers can reason on the trade-off between path safety and efficiency. 

In this work, we improve our previous work on stochastic functional gradient planners. We introduce a novel expressive path representation based on kernel approximation, that allows cost effective model updates based on stochastic samples. The main drawback of the previous stochastic functional gradient planner was the cubic cost, stemming from its non-parametric path representation. Our novel approximate kernel based model, on the other hand, has a fixed linear cost that depends solely on the number of features used to represent the path. We show that the stochasticity of the samples is crucial for the planner and present comparisons to other state-of-the-art planning methods in both simulation and with real occupancy data. The experiments demonstrate the advantages of the stochastic approximate kernel method for path planning in occupancy maps.   

\end{abstract}

\section{Introduction} \label{sec:intro}

Path planning is a critical decision making process in autonomous robotics. Its foremost requirement is path safety, which guarantees an obstacle free motion from the robot's current configuration to its goal. As this is a pivotal aspect of autonomy, path planning has been a long studied subject of robotics with a prolific range of planning methods. While the exact method changes between planners, the mechanism to find a safe path typically takes the form of either a sampling-based approach or of a trajectory optimiser. 
  
Occupancy maps are a probabilistic representation of the robot's environment as it is perceived from noisy sensor observations \cite{Thrun2009}. A map is a discriminative model, which returns the probability that an obstacle is present. Planning on occupancy maps is most commonly done by sampling-based planners \cite{Tsardoulias2016}. These methods are probabilistically guaranteed to find a safe path but are not explicitly optimising any objective function such as path length or execution time. To alleviate this problem, most sampling-based planners employ a second heuristic-based step to improve the initial solution. 

Trajectory optimisers offer a different take on path planning using a variational approach. This enables optimisation of any objective function, such as safety or control cost, directly in the space of trajectories. However, aside from \cite{Francis2017}, there are no implementations of trajectory optimisers for occupancy maps. The main impediment lies in the optimiser's main assumption that the objective cost and gradient can be computed anywhere. Such a general assumption is not applicable in occupancy maps, as the map might have gaps or non-informative gradients. Consequently, there are no guarantees for an optimal or even safe solution.

In this paper, we present a novel approach for trajectory optimisation using occupancy maps. We utilise kernel approximation techniques to form an expressive and tractable non-linear path model. Aside from its low computational requirements, the path model is naturally updated by stochastic samples. Other functional gradient path planning techniques (e.g. \cite{Marinho2016}, \cite{mukadam2016gaussian}) employ a predetermined sampling resolution, which might be insufficient for planning in occupancy maps. The proposed planner, on the other hand, does not commit to any sampling resolution. Rather, it draw samples from the entire domain, which ensures that the model can  react anywhere along the curve.

The technical contributions of this paper are:
\begin{enumerate}
	\item An expressive and tractable path model based on kernel approximations, which can be considered as a generalisation of the Gaussian Process Motion Planner \cite{mukadam2016gaussian}. This is a critical building block of the path planner, as it is allows fast and low computational cost update procedures using stochastic samples.
	\item Employing \textit{Stochastic Gradient Descent} (SGD) \cite{Bottou2010} in the path planning paradigm to ensure convergence to an optimal solution under the guarantees of SGD. SGD exploits kernel approximation methods in order to keep the model tractable, unlike the non-parametric approach taken in \cite{Francis2017}. 
 \end{enumerate}

The remainder of this paper is organised as follows. Section \ref{sec:RelatedWork} reviews the literature on path planning using occupancy maps. Section \ref{sec:preliminaries} provides background on functional gradient methods and their adaptation for path planning. Section \ref{sec:methods} provides details on the core elements of the proposed method. The results obtained in various simulation and real data scenarios are shown in Section~\ref{sec:results}. Finally, Section~\ref{sec:conclusions} draws conclusions about the proposed method.

\section{Related Work} \label{sec:RelatedWork}

Sampling-based methods have been predominately used for path planning in occupancy maps with a wide range of successful algorithms such as 
\textit{Rapidly exploring Random Trees} (RRT), \textit{Probabilistic Road Map} (PRM) and Space skeletonisation. A comparison of the performance of these methods for planning in occupancy maps is described in \cite{Tsardoulias2016}. Working in the robot's configuration space, these methods typically break the planning methods into two steps. First, a graph-like representation of the configuration space is created from samples. In this data structure, the vertices and edges represent safe configurations and connections, respectively. The second stage finds a path using a search algorithm on the graph structure. During that stage a heuristic-based criteria can be employed to improve the resulting path characteristics. Space skeletonisation methods build a one-dimensional skeleton of the configuration space using a variety of methods such as visibility graphs and Voronoi diagrams, which is then used to compute safe paths \cite{lozano1979algorithm, bhattacharya2007voronoi, garrido2006path}. PRMs are another instance of the skeletonisation methods, where vertices and edges are randomly sampled from the configuration space \cite{Kavraki1996}. A tree search method, such as Dijkstra or A$^*$, is used to compute the resulting path. RRTs are a family of efficient algorithms for searching high-dimensional spaces proposed by LaValle \cite{Lavalle98rapidly-exploringrandom}. RRTs grow a tree from the starting pose by randomly adding new nodes. While sampling-based methods are highly successful in obtaining safe paths using occupancy maps, it is clear that these are suboptimal solutions as there is no explicit optimisation criteria applied during the sampling process.

Another path planning approach, although not commonly used with occupancy maps, is trajectory optimisation. In this planning paradigm the resulting path is a stationary solution of an explicit optimisation problem defined by a cost function. The cost function provides a measure of path optimality which can arise from a variety of criteria, e.g. distance from obstacles or motion costs. Khatib \cite{Khatib1986} introduced a path planning method based on artificial potential fields. Zucker et al. \cite{Zucker2013}, in their work on \textit{Covariant Hamiltonian Optimisation for Motion Planning} (CHOMP), reframed path optimisation as a variational problem, where path optimisation is performed directly in the space of trajectories. \textit{Stochastic Trajectory Optimisation for Motion Planning} (STOMP) \cite{Kalakrishnan2011} performs optimisation by exploring the space of trajectories using noisy perturbations. The cost functional in STOMP can be non-differentiable, as the iterative update rule utilises an auxiliary gradient function based on the stochastic samples. The main limitation of both CHOMP and STOMP is the waypoint parameterisation used to represent the solution. Using such a path representation, although very intuitive, requires a good balance between path expressiveness and computational complexity. In recent years, several methods introduced optimisation on smooth paths instead of the discrete waypoint representations. Mukadam et al. \cite{mukadam2016gaussian} employed \textit{Gaussian Processes (GP)} generated by linear time varying stochastic differential equations to represent a path. Dong et al. \cite{Dong2016} extended this work by reformulating optimisation as a probabilistic inference problem where the path is the \textit{maximum a posteriori} (MAP) solution. Marinho et al. \cite{Marinho2016} used \textit{Reproducing Kernel Hilbert Space} (RKHS) to form a non-parametric path representation. However, all methods fail to generate a safe path while planning in occupancy maps as discussed in section \ref{sec:methods}.

\section{Preliminaries} \label{sec:preliminaries}

\subsection{Functional Gradient Descent Optimisation} \label{subec:FGDO}
In this section, we describe the theory of functional gradient optimisation methods and their application in path planning. \textit{Functional gradient descent} (FGD) forms a variational framework for optimising costs. However, in the context of path planning, their main objective is to produce a safe, collision-free path. A secondary objective may incorporate other costs such as smoothness of the trajectory or time of travel.

We begin by first introducing notation. A path, $\xi: [0,1] \rightarrow \mathcal{C}\in\mathbb{R}^D$, is a function that maps time-like parameter $t \in [0,1]$ into configuration space $\mathcal{C}$. We define an objective functional $\mathcal{U}(\xi): \Xi \rightarrow \mathbb{R}$ that returns a real number for each path $\xi \in \Xi$. The objective functional is used in the optimisation process to capture the path optimisation criteria such as smoothness and safety. 

Regardless of the exact choice of cost functional $\mathcal{U}(\xi)$, $\xi$ can be optimised by following the functional gradient. Similar to other gradient descent methods, optimisation is performed iteratively. The functional gradient update rule is derived from a linear approximation of the cost functional around the current trajectory, $\xi_n$:
\begin{equation}
	\mathcal{U}(\xi) \approx \mathcal{U}(\xi_n)+\nabla_\xi\mathcal{U}(\xi_n)(\xi-\xi_n) + \mathcal{O}((\xi-\xi_n)^2).
\end{equation}
To ensure convexity of the objective function, we add a regularisation term based on the norm of the update:
\begin{equation}\label{eq:FGMP_optimisaton}
	\xi_{n+1}= \argmin_\xi \quad \mathcal{U}(\xi_n) + (\xi-\xi_n)^T\nabla_\xi\mathcal{U}(\xi_n) + \frac{1}{2\eta_n}\|\xi-\xi_n\|_M^2.
\end{equation}
The regularisation term $\|\xi-\xi_n\|_M^2 = (\xi-\xi_n)^TM(\xi-\xi_n)$ is the squared norm with respect to a metric tensor $M$ and $\eta_n$ is a user-defined learning rate. A closed form solution of (\ref{eq:FGMP_optimisaton}) is derived by differentiating the right hand side of (\ref{eq:FGMP_optimisaton}) with respect to $\xi$ and setting it to zero, yielding:
\begin{equation}\label{eq:FGMP_update_rule}
	\xi_{n+1}(\cdot)= \xi_n(\cdot) - \eta_n M^{-1}\nabla_\xi\mathcal{U}(\xi_n).
\end{equation}
The form of the update rule in (\ref{eq:FGMP_update_rule}) is general, and thus, invariant to the choice of the objective function or the solution space representation. The only requirements are that $M$ is invertible and the gradient, $\nabla_\xi\mathcal{U}(\xi_n)$ exists.

\subsection{FGD for Motion Planning} \label{subsec:FGMP}

FGD establishes a general framework for optimisation. To apply it in a motion planning context, an objective functional must be specified.  

The objective functional $\mathcal{U}(\xi)$ in a motion planner paradigm consists typically of a weighted sum of at least two penalties; \begin{inparaenum}[(i)]  
 	\item $\mathcal{U}_{obs}(\xi)$ which encodes a penalty based on proximity to obstacles;
 	\item $\mathcal{U}_{dyn}(\xi)$ that regulates and constrains the curve shape or space-time dynamics. Combining both penalties using a user-defined regularization coefficient $\lambda$ we obtain the objective function:
 \end{inparaenum}   
\begin{equation}\label{eq:FPMP_U}
\mathcal{U}(\xi) = \mathcal{U}_{obs}(\xi) + \lambda\mathcal{U}_{dyn}(\xi).
\end{equation}

In the following sections we define the objective functionals, $\mathcal{U}_{obs}$ and $\mathcal{U}_{dyn}$,  and their corresponding functional gradients.

\subsubsection{Obstacle Functional $\mathcal{U}_{obs}(\xi)$} \label{subsubsec:obs_fnctl}

Obstacles lie in the robot's working space $\mathcal{W} \in \mathbb{R}^3$. However, the path $\xi$ is defined in configuration space $\mathcal{C}$. Hence, estimating the obstacle cost functional requires mapping of $\xi(t)$ from configuration space into workspace using a forward kinematic map $x$. To account for the size of the robot or any uncertainty in its pose, a set of body points on the robot, $\mathcal{B} \in \mathbb{R}^3$ is defined. $x\left(\xi(t),u\right)$ maps a robot configuration $\xi(t)$ and body point $u \in \mathcal{B}$ to a point in the workspace $x: \mathcal{C} \times \mathcal{B} \rightarrow \mathcal{W}$. As the obstacle functional returns a single value for each $\xi(\cdot)$, its return value is calculated by aggregating the workspace cost function, $c: \mathbb{R}^3 \rightarrow \mathbb{R}$, along the trajectory and robot body points using a \textit{reduce} operator. An example for such an operator can be an integral or a maximum. We require that the \textit{reduce} operator would be approximately represented by a sum over a finite set, $\mathcal{T}(\xi) = \{t,u\}_i$ of time and body points:
\begin{equation}\label{eq:FGMP_Uobs}
\mathcal{U}_{obs}(\xi) \approx \sum_{(t,u)\in\mathcal{T}(\xi)} c\left(x\left(\xi(t),u\right)\right).
\end{equation}

\subsubsection{Path Dynamics Functional $\mathcal{U}_{dyn}(\xi)$} \label{subsubsec:dyn_fnctl}

$\mathcal{U}_{dyn}(\xi)$ acts as a regularisation term that penalises based on the kinematic costs associated with $\xi$. A common approach is to penalise on the trajectory length by optimising the integral over the squared velocity norm:
\begin{equation}\label{eq:velocity_norm}
	\mathcal{U}_{dyn}(\xi) = \frac{1}{2}\int_{0}^{1} \left|\left|\frac{d}{dt}\xi(t) \right|\right|^2 dt.
\end{equation}  
Other methods, such as in \cite{Marinho2016}, regularise using the $L_2$ norm of $\xi$. Although this is a straight-forward and simple regularisation, it is less desirable for general path planning tasks. The use of $L_2$ norm as a regularisation term asserts implicitly the zero-line, connecting the start and goal points, as a preferred solution. Instead, it is better to use (\ref{eq:velocity_norm}) as a regulariser and define explicitly a mean function to bias the FGD solution.

\subsubsection{Functional Gradients} \label{subsubsec:func_grad}

To implement the iterative update rule of FGD, the functional gradient of  $\mathcal{U}_{obs}$ and  $\mathcal{U}_{dyn}$ must be defined. As both objective functionals are of the form $\mathcal{F}(\xi)=\int_{a}^{b} v(t,\xi,\xi') dt$, we can write the functional gradient as \cite{Zucker2013}:
\begin{equation}\label{eq:functional_grad}
	\nabla\mathcal{F}_\xi(\xi)=\frac{\partial v}{\partial\xi} - \frac{d}{dt}\frac{\partial v}{\partial\xi'}.
\end{equation}
Eq. \ref{eq:FPMP_U} defines the objective functional as a weighted sum of separate penalties. Therefore the functional gradient can be computed as the sum of the different gradient terms:
\begin{equation}\label{eq:FPMP_dU_dxi}
\nabla_\xi\mathcal{U}(\xi) = \nabla_\xi\mathcal{U}_{obs}(\xi) + \lambda\nabla_\xi\mathcal{U}_{dyn}(\xi).
\end{equation}

Applying (\ref{eq:functional_grad}) to the obstacle objective $\mathcal{U}_{obs}$ yields;
\begin{equation}\label{eq:d_U_obs-final}
\nabla_\xi\mathcal{U}_{obs}(\xi(t),u)=\frac{\partial}{\partial \xi(t)}x(\xi(t),u) \nabla_x c\left(x\left(\xi(t),u\right)\right),
\end{equation}
where $\boldsymbol{J}(t,u) \equiv \frac{\partial}{\partial \xi(t)}x(\xi(t),u)$ is the workspace Jacobian and $\nabla_x$ is the Euclidean gradient of the cost function $c$. With (\ref{eq:velocity_norm}) as choice for the dynamic penalty $\mathcal{U}_{dyn}$ the functional gradient can be easily computed using (\ref{eq:functional_grad}) as:
\begin{equation}\label{eq:d_U_reg-final}
	\nabla_\xi\mathcal{U}_{dyn}(\xi(t))= -\frac{d^2}{dt^2}\xi(t).
\end{equation}

\section{Methods} \label{sec:methods}

FGD is an efficient method for path optimisation. However, the current implementations are not suitable for planning with occupancy maps. We identify two main reasons which are discussed and resolved in the following sections. First, the map's obstacle functional and its spatial gradient have a counter intuitive form, which differ significantly from the well-behaved obstacle functional used by other planners (e.g. \cite{Zucker2013,Marinho2016,mukadam2016gaussian}). Second, sampling of the objective function, and as a result, the path representation follow a deterministic scheme. Therefore, such planners lack any guarantees for convergence of the solution to a safe path. Our approach uses a stochastic gradient update rule combined with an approximate kernel path representation to ensure sampling along the entire curve whilst keeping a closed-form and concise path model.

\subsection{Occupancy Map Obstacle Functional} \label{subsec:OC_func}

Most FGD motion planners precompute the obstacle cost $c$ and its spatial gradient based on the distance to known obstacles (e.g \cite{Zucker2013,Marinho2016}). This approach allows for a fast and cheap retrieval of gradients during the optimisation. However, in many autonomous planning scenarios the robot has only limited knowledge of obstacle properties such as location and size. Sensors provide probabilistic information about the location of obstacle borders. Yet, contextual data about the obstacles are not easily inferred. In addition, using a discretised space representation prohibits the use of continuous mapping methods such as the Gaussian Process Occupancy Maps \cite{OCallaghan2012} and Hilbert maps \cite{ramos2015hilbert}.   

\begin{figure}[!ht]
	\centering
    \begin{subfigure}[b]{0.22\textwidth}
        \includegraphics[width=\textwidth]{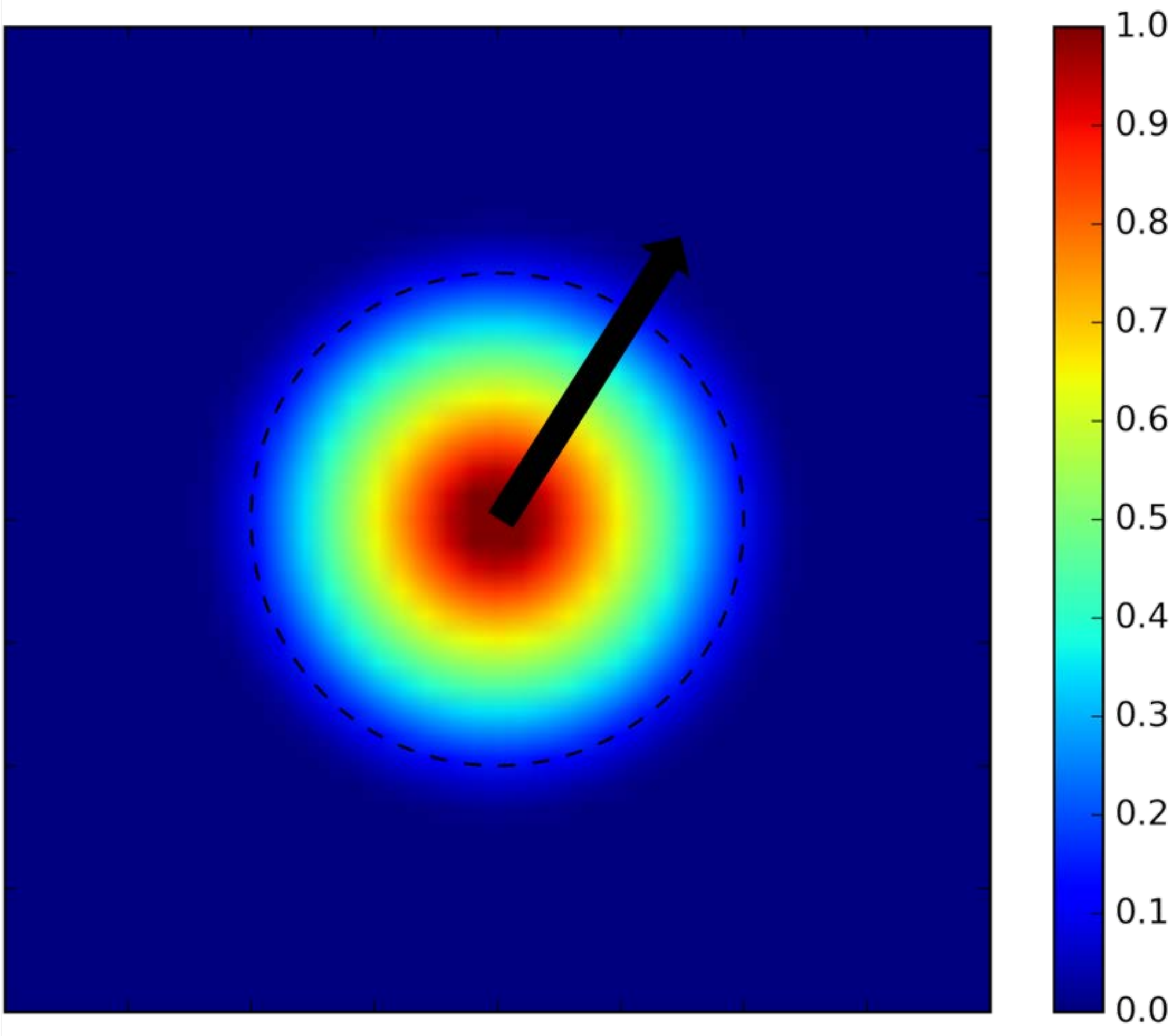}
        \caption{Precomputed cost map}
        \label{costmap:subfig-1}
    \end{subfigure}
    \hfill
    \begin{subfigure}[b]{0.22\textwidth}
        \includegraphics[width=\textwidth]{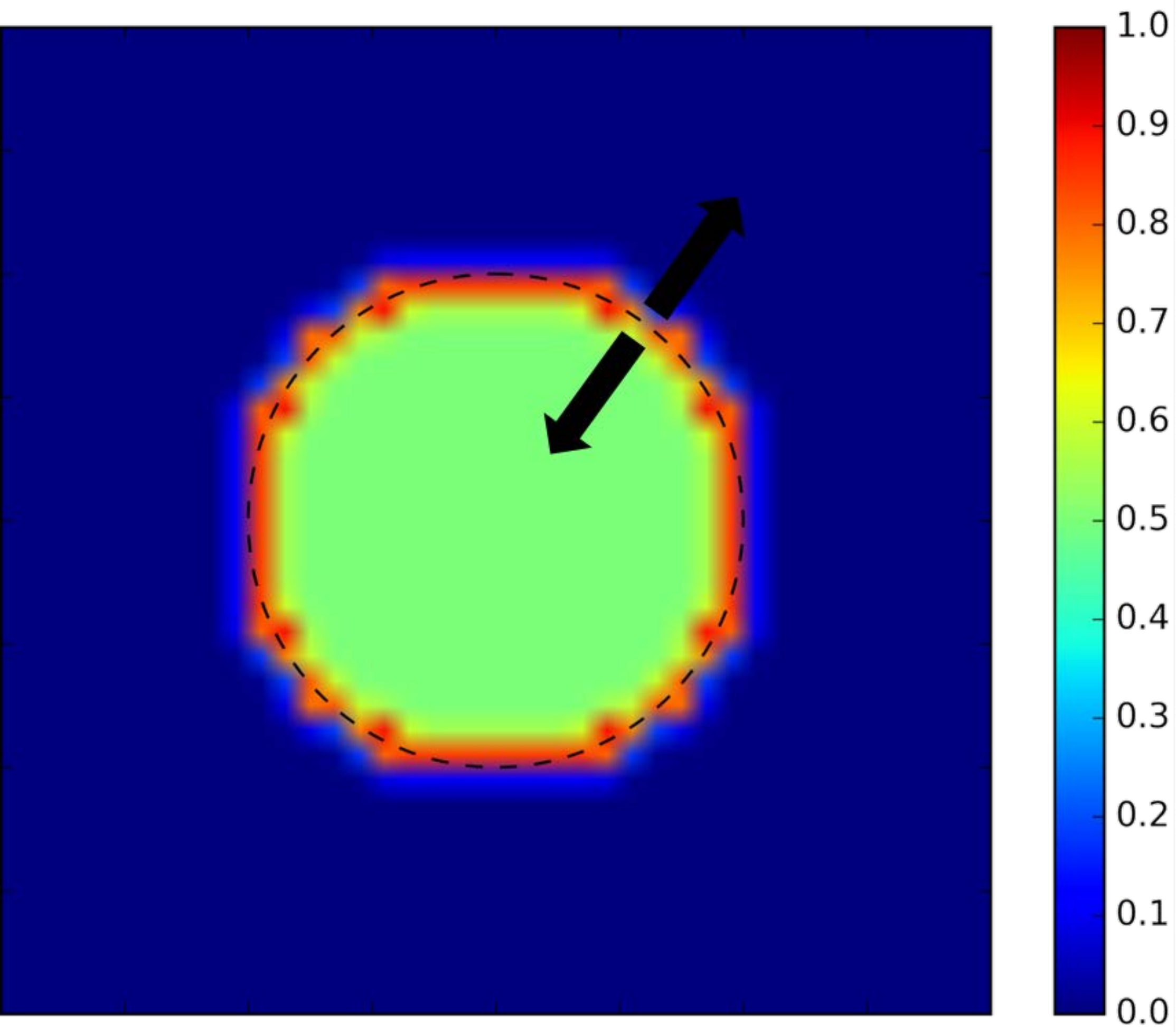}
        \caption{Occupancy Grid Map}
        \label{costmap:subfig-2}
    \end{subfigure}
	\par \bigskip
     \begin{subfigure}[b]{0.25\textwidth}
        \includegraphics[width=\textwidth]{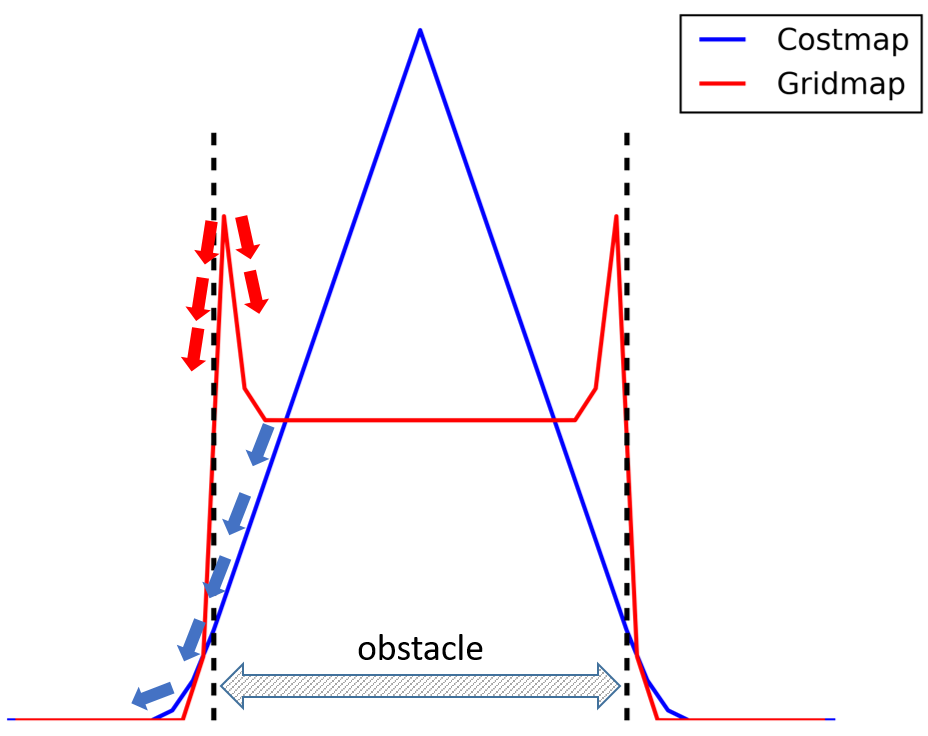}
        \caption{Cross section}
        \label{costmap:subfig-3}
    \end{subfigure}
    \par \smallskip
 	\caption{Comparison of cost maps used in path planning. Dashed lines illustrate the obstacle boundaries. Arrows indicates the direction of gradient. \protect\subref{costmap:subfig-1} precomputed cost field as specified in \cite{Zucker2013}. \protect\subref{costmap:subfig-2} shows an occupancy map based on sensors observations.\protect\subref{costmap:subfig-3} depicts cross sections of the map in \ref{costmap:subfig-1} and \ref{costmap:subfig-2}, which emphasise the difference in spatial gradients, indicated by corresponding coloured arrows. In the cost map, cost is defined everywhere in the map and always generates repulsive gradient. The direction of gradient in occupancy map is inconsistent around the obstacle borders, with attractive potential in unsafe areas of the map.}
	\label{fig:costmap}
\end{figure}

Fig. \ref{fig:costmap} illustrates the differences between the precomputed cost used by most FGD motion planners and a standard occupancy map \cite{Elfes1989}. Fig. \ref{costmap:subfig-1} shows a cost map with complete knowledge of the obstacle. The cost, given in closed-form, and its spatial gradient are defined everywhere in the map as indicated schematically by the arrows. Fig. \ref{costmap:subfig-2} illustrates the equivalent occupancy grid map, which is inferred from laser observations. As expected, The grid map only holds information about observed locations. In those regions, the cost follows the occupancy. However, in the unobserved regions of the map, the two approaches generate different outputs. While the precomputed cost still produces a well-behaved function and the desired repulsive gradient, the occupancy returns to the map's prior occupancy probability of 0.5 and generates inconsistent gradients. 

The main challenge of using FGD on occupancy maps lies in the inability to define a usable gradient everywhere in the map. Cross sectional data of the precomputed cost and occupancy maps, as depicted in Fig. \ref{costmap:subfig-3}, summarises this. In observed areas, the occupancy can act as the obstacle cost, as spatial gradients of both maps "pushes" away from obstacles. However, in occluded regions the behaviour is entirely different. While the precomputed gradient still returns a repulsive gradient, pushing away from the obstacle, the spatial gradient of the occupancy map pulls inward, toward unobserved and unsafe regions of the map.

The approach taken in this work relies on the observed occupancy of each sample point to decide whether to accept or reject a gradient update. How to obtain the spatial gradient $\nabla_x c\left(x\left(\xi(t),u\right)\right)$ depends on the mapping method used. In an occupancy grid map, the gradient can be approximated from one of several gradient operators used in computer vision, e.g. Sobel-Feldman or Canny operators \cite{Kaehler2008}. We note again that there is no need to precompute any cost or gradient, as these are estimated on-line where it is required. In this work, however, we opted to work with Hilbert maps \cite{ramos2015hilbert}, which provide a fast and continuous occupancy map model.

We follow our previous work, presented in \cite{Francis2017}, to compute spatial occupancy gradient directly from the map model. A Hilbert map is a discriminative probabilistic model;
\begin{equation} \label{eq:Hmap_prob}
 	p(y^* = +1|\boldsymbol{x}^*),
\end{equation}
that returns the probability of occupancy for any query point $\boldsymbol{x}^*$. As the model is continuous and at least twice differentiable \cite{ramos2015hilbert}, we can compute in closed-form the spatial gradient of the probability of occupancy and assign it to $\nabla_x c\left(x\left(\xi(t),u\right)\right)$:
\begin{equation}
\nabla_x c\left(x\left(\xi(t),u\right)\right) = \frac{\partial}{\partial\boldsymbol{x}^*}p(y^* = +1|\boldsymbol{x}^*).
\end{equation}

\subsection{Stochastic Functional Regression}\label{subsec:SFR}

Any FGD planner optimises an objective function, such as in (\ref{eq:FPMP_U}). As the objective is uncountable, it is estimated via samples. Therefore, the choice of sampling schedule is cardinal for a successful and efficient planner. The importance of the sampling schedule is exacerbated in occupancy maps, where not every sample can generate an informative gradient. Consequently, sampling everywhere along the curve is most desired, as this increases the chance of identifying transition areas in the map. Yet, with a fixed resolution sampling defining a sufficient resolution a-priori is difficult. Hence most methods limit the sampling resolution according to their computational resources.

GP-based planners \cite{mukadam2016gaussian,Dong2016} use GPs for a smooth path representation. However, as the path is updated only at the support points, it requires a dense representation in order to ensure sufficient expressivity. Similar limitations also hold for the non-parametric approach used in \cite{Marinho2016}, where the support is taken from fixed resolution samples of the objective function. CHOMP and STOMP perform batch optimisation by exploring the solution space using either Hamiltonian Monte Carlo or by estimating the probability density of the objective using noisy path perturbations. As the path is waypoint based the solution space exploration is performed in the robot's workspace. Consequently, the optimisation process is highly sensitive to the choice of the exploration hyperparameters. For example, STOMP's update rule fails if the variance of perturbation is smaller than the size of obstacles, which in occupancy maps is unknown a-priori. The stochastic non-parametric approach of \cite{Francis2017} addresses this problem by using continuous sampling in the trajectory domain. However, as path is represented by a GP the computational costs are high, i.e. of the order $\mathcal{O}(N^3)$ where N is the number of samples.

The approach taken in this work, alleviates the limitations present in previous work. Namely, it allows stochastic updates from continuous samples. To keep the computational cost low while maintaining a highly expressive representation, a parametric and thus concise path representation based on kernel approximation is employed.  

In the approximate kernel approach we denote $\hat{\Upsilon}(t')$ as a finite set of features consisting of $t' \in [0,1]$. The choice of features in a set is dictated by the desired kernel, with the aim of approximating the \textit{Reproducing Kernel Hilbert Space} (RKHS) inner product, $\langle\cdot,\cdot\rangle$;
\begin{equation} \label{eq:kernel_approx}
	k(t,t') = \langle\Upsilon(t),\Upsilon(t')\rangle \approx \hat{\Upsilon}(t)^T\hat{\Upsilon}(t').
\end{equation} We note that the set of features only approximates the selected kernel in expectation, hence the $\hat{}$ notation. Using a weight vector $\mathbf{w}$ we can now express the robot configuration at $t$, $\xi(t)$, as a function of a the finite set of approximating features, $\hat{\Upsilon}(t')$:
\begin{equation}\label{eq:approx_krnl_path}
	\xi(t) = \xi_{o}(t) + \xi_{b}(t) + \mathbf{w}^T\langle\hat{\Upsilon}(t'),\hat{\Upsilon}(t)\rangle.
\end{equation}
$\xi_o$ is an offset path, which may be used to bias solution and can be computed by a crude and fast planner. $\xi_b$ is a term used to adjust boundary conditions. Both $\xi_o$ and $\xi_b$ are represented by an approximated kernel representation with the same curve properties as $\xi$ (continuity, derivability, etc.), although the feature set might be different. The approximating features $\hat{\Upsilon}(\cdot)$ can take different forms. Common kernel matrix approximation are \textit{Random Fourier Features} (RFF) \cite{Rahimi2008} and the Nystr\"{o}m approximation \cite{williams2001using}.

Once the path representation has been defined, we can treat path planning as a regression problem, i.e., optimising the weight vector $\mathbf{w}$:
\begin{equation}
	\mathbf{w}_{optimal}= \argmin_\mathbf{w} \quad \mathcal{U}(\mathbf{w}).
\end{equation} The advantage of using this approach is that the model can be learned through stochastic sequential updates from continuous samples. 

In the following sections we discuss how to implement FGD using the approximated kernel regression model. We revise the general update rule of (\ref{eq:FGMP_update_rule}) into a practical gradient update based on the choice of path representation. Then, we lay out the full algorithm of the stochastic approximate kernel path planner.

\subsection{Approximate Kernel Update Rule } \label{subsec:update_rule}
Using approximated kernels keeps the path representation both smooth and concise. However, in this section we discuss how the general update rule of \ref{eq:FGMP_update_rule} is implemented in practice.

Eq. (\ref{eq:approx_krnl_path}) expresses the path as a weighted sum of features. Therefore the iterative update rule of (\ref{eq:FGMP_update_rule}) must be performed with respect to the weight vector $\mathbf{w}$. Following (\ref{eq:FGMP_update_rule}), we sample the functional gradient of the objective function at time $t_i$. We refer to these samples as stochastic, since $t_i$ can be drawn at random from anywhere along the curve domain, $t_i \in [0,1]$, and is not limited by a predefined sampling resolution. The sampled gradient $g(t_i)$ can be viewed as a path perturbation $g(t_i) = \nabla_\xi\mathcal{U}(\xi_n)(t_i)\Upsilon(t_i)$. As $g$ is defined in the full RKHS of $k$, it must be projected onto the solution space spanned by $\mathbf{w}$ using the appropriate inner product, which can be approximated using (\ref{eq:kernel_approx}) as:
\begin{equation} \label{eq:update_rule_weights}
	\begin{split}
	\mathbf{w}_{n+1} &= \mathbf{w}_n - \eta_n M^{-1} \langle\Upsilon(t'),	\nabla_\xi\mathcal{U}(\xi_n)(t_i)\Upsilon(t_i)\rangle \\ 
    &= \mathbf{w}_n - \eta_n M^{-1} \langle\Upsilon(t'),\Upsilon(t_i)\rangle\nabla_\xi\mathcal{U}(\xi_n)(t_i)\\
	&\approx \mathbf{w}_n - \eta_n M^{-1} \hat{\Upsilon}(t')^T\hat{\Upsilon}(t_i)\nabla_\xi\mathcal{U}(\xi_n)(t_i).
	\end{split}
\end{equation}
Note that to guarantee convergence of SGD, the learning rate $\eta_n$ must satisfy the Robbins-Monro conditions \cite{Robbins1951};   $\sum_{n=1} \eta_{n}^2 < \infty$ and $\sum_{n=1} \eta_{n} =  \infty$.

Boundary conditions are handled in a similar fashion. We employ an additional path $\xi_{b} = \mathbf{w}_b^T\hat{\Upsilon_b}$ to compensate the boundary conditions. The boundary features $\hat{\Upsilon_b}$ are not necessarily identical to $\hat{\Upsilon}$. The update rule for $\mathbf{w}_b$:
\begin{equation}\label{eq:boundary}
	{\mathbf{w}_b}_{n+1} = {\mathbf{w}_b}_n -  M^{-1} \hat{\Upsilon_b}(t_b)^T\hat{\Upsilon_b}(t_b) \Delta x_b(t_b).
\end{equation}
Here $t_b$ are time points were boundary conditions are defined and $\Delta x_b(t_b)$ is the corresponding difference between the current value of of $\xi$ at $t_b$ and the desired value at the boundary. Note that this similar to  (\ref{eq:update_rule_weights}), except $\eta_n$ was omitted and the gradient was replaced by the difference to the desired boundary value. 

\subsection{Path Planning Algorithm} \label{subsec:algo}

The pseudo-code of the stochastic approximate kernel path planner is shown in Algorithm (\ref{algo:main}). The output of this algorithm is an optimised path $\xi_{min}(\cdot)$, parametrised by the weight vector $\mathbf{w}_{min}$.

At each iteration, a mini-batch $(t_{s}, u_{s})$ is drawn uniformly. The occupancy of each sample $t^* \in t_{s}$ and $u^* \in u_{s}$ is assessed by querying the map model in the corresponding state $\xi_{n}(t^*)$. If the probability of occupancy at $\xi_{n}(t^*)$, $P_{occ}$, is within the safety limits, i.e. clear of obstacles, a functional gradient update is invoked. Following (\ref{eq:update_rule_weights}), the weight vector $\mathbf{w}$ is updated with the stochastically sampled gradient observations, leading to a new path representation $\xi_{n+1}(\cdot)$. Finally, the boundary condition are enforced using (\ref{eq:boundary}). 

The low computational complexity of this algorithm stems from the concise path representation and update rule. Using $M$ approximated features, the computational cost of updating and querying the path model is linear and fixed as $\mathcal{O}(M)$. This is in contrast to the computational cost of the stochastic GP path planner which is cubic with the number of updates. 

\begin{algorithm}[bt]
	\caption{Stochastic approximate kernel FGD path planner}
	\label{algo:main}
	\DontPrintSemicolon
	\KwIn{$\mathcal{H}$: Occupancy Map.}
	\myinput{$\xi(0),\xi(1)$: Start and Goal states.}
	\myinput{$P_{safe}$: No obstacle threshold.}
	\myinput{${\hat{\Upsilon(t',\cdot)}}$: Approximated feature vector.}
    
	\KwOut{$\mathbf{w}_{min}, \xi_{min}(\cdot)$}

	{$ $}\\
	$n=0$\\
	\While{solution not converged }
	{
		Stochastic sampling:	\\
		$(t_{s},u_{s}) \sim \mathcal{U}[0,1] \leftarrow$ Draw mini-batch uniformly \\
		\ForEach{$(t^*,u^*) \in (t_{s},u_{s})$}
		{
        $P_{occ} \leftarrow \mathcal{H}(x(\xi_n(t^*),u^*))$ Eq. \ref{eq:Hmap_prob}\\
			\If{$P_{occ} \leq P_{Safe}$}
            {
            $\mathbf{w}_{n+1} \leftarrow$ update rule Eq. \ref{eq:update_rule_weights}\\
            }
      Fix boundary conditions, Eq. (\ref{eq:boundary}) \\
        }
       
		$n=n+1$	
	}
\end{algorithm} 

\section{Results} \label{sec:results}

In this section, we evaluate the performance of our method and compare it with other related path planning techniques in simulation and with real data. We show that stochastic sampling is a critical aspect of path planning in occupancy maps, which is complimented by the scalable model of the the approximate kernel path representation.

\subsection{Simulation}

Most trajectory optimisers assume full knowledge of obstacle properties and compute a cost function and its spatial gradient for the entire workspace, e.g. \cite{Zucker2013}. This is not attainable when working with occupancy maps. 

\begin{figure}[thpb]
	
	\centering
	
	\includegraphics[width=0.48\textwidth]{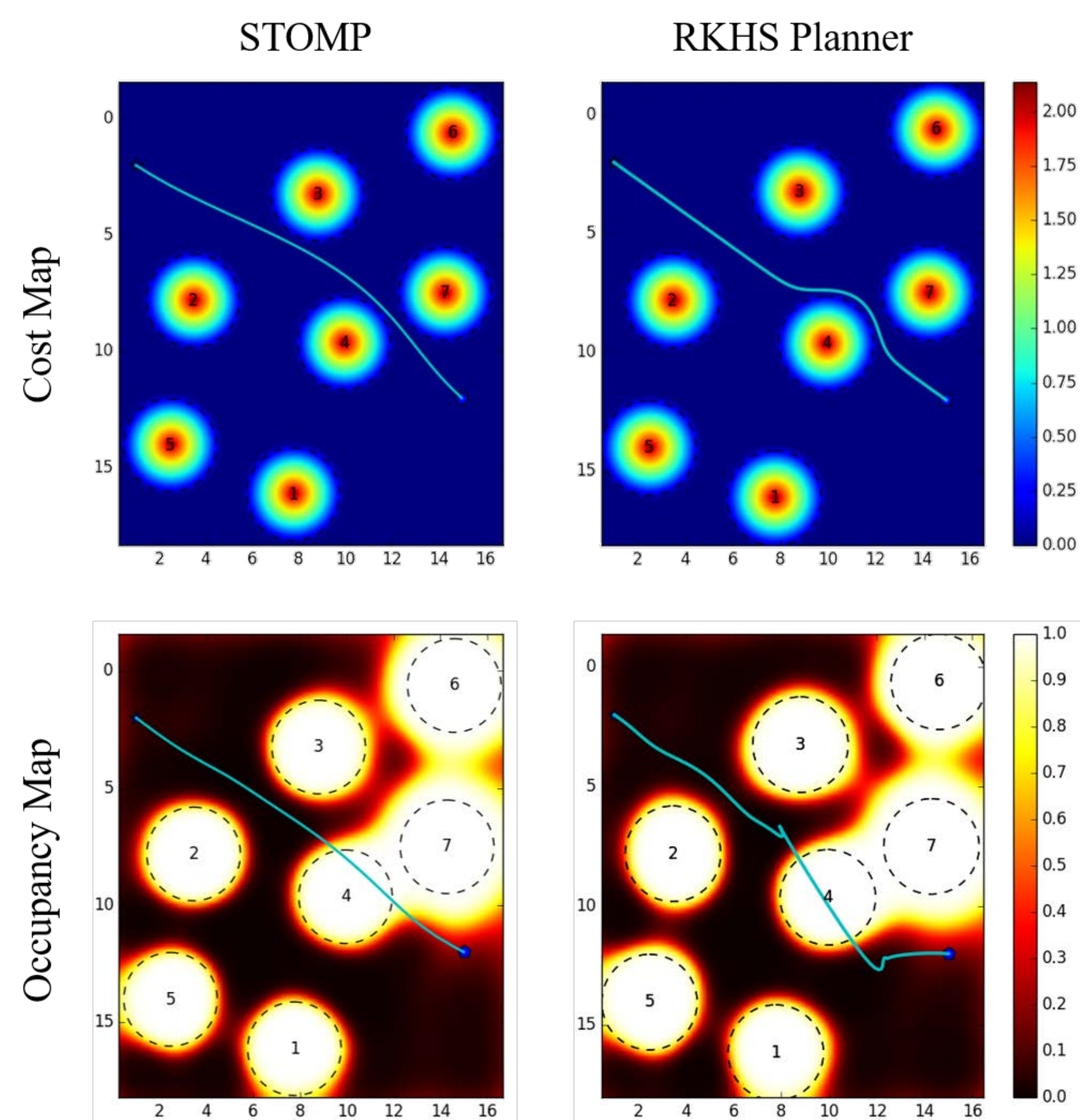}
	
	\caption{Comparison of motion planning in cost maps vs. occupancy maps using state-of-the-art planners. In all maps, dashed black lines indicate the borders of obstacles. Cost map is computed according to \cite{Zucker2013}. Hilbert maps \cite{ramos2015hilbert} are used to model occupancy, which is computed from simulated laser readings. Left, STOMP \cite{Kalakrishnan2011} and on the right, RKHS non-parametric planner \cite{Marinho2016}. Both planners are successful when planning in a well-defined cost map. However, both fail when planning using occupancy maps.} 
	\label{fig:comapre_methods}
    \hspace{-5mm}
\end{figure}

Fig. \ref{fig:comapre_methods} compares planning using cost maps and occupancy map using two leading methods, STOMP \cite{Kalakrishnan2011} and an RKHS non-parametric planner \cite{Marinho2016}. Occupancy is represented by a Hilbert map \cite{ramos2015hilbert}, which was computed using simulated laser readings of the environment. While planning in a cost map both methods successfully find a safe path from start to goal. However, when the same algorithms are used with the occupancy map, they both fail. The reason lies in the deterministic sampling schedule both methods use, where path is updated only around its predetermined support. As there are no valid gradients inside the obstacles, gaps are formed in the support of both curves and the path can not be updated.

\begin{figure*}[bt]
 \centering
 \includegraphics[width=0.9\textwidth]{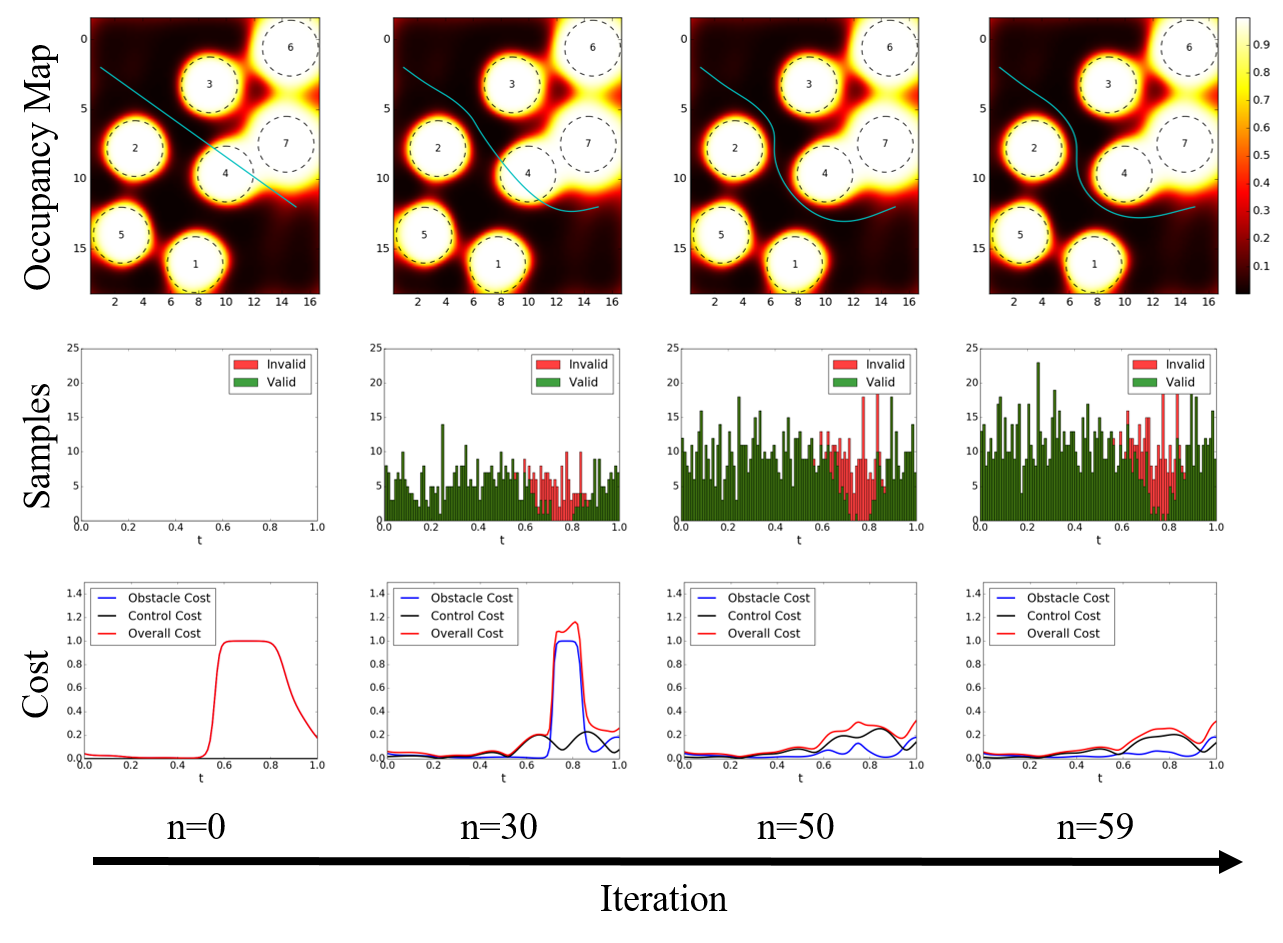}
        	\caption{The image is divided into four segments, depicting the state of our planner at different iterations $n=0,30,50,59$. Each segment's column consists of three images: (top) shows the current path superimposed on the occupancy map, (middle) shows the accumulated samples over all previous iterations, where green and red indicate valid and invalid samples respectively, (bottom) depicts the objective functions for obstacle, motion-related and the overall cost. This view of the cost is only for presentation purposes and is not available to the planner, as it relies only on stochastic samples.  $n=0$: The planner starts from an initial guess, $\xi_o$, which in this example is the line connecting the start and goal points. $n=30$: The path deforms around the edges of the obstacles, while samples are drawn across the entire domain $[0,1]$. $n=50$: Overall cost has reduced and path clears all obstacles. $n=59$: Solution had converged.}
\label{fig:LinReg_Nystroem} 
 \end{figure*}

The performance of our proposed stochastic planner differs from that of other planners. An overviews of the iterative process of the stochastic FGD path planner is shown in Fig. \ref{fig:LinReg_Nystroem} where each column shows \begin{inparaenum}[(i)]  
 	\item the current path overlaid on the occupancy map;
 	\item the accumulated samples, both valid and invalid;
    \item the underlying cos which is used for presentation purposes only as it is available to the planner only through stochastic samples.
 \end{inparaenum} 
The iterative update process starts from an initial guess, $\xi_o$. In Fig. \ref{fig:LinReg_Nystroem}, the planner initialises with the line connecting the start and goal points. The overall cost consists only on the obstacle cost as the dynamic cost for a straight line are $\mathcal{U}_{dyn} = 0$. After 30 iterations path deforms around the edges of the obstacles. Samples are drawn from the entire domain $[0,1]$. Samples inside the obstacles are rejected. However, samples on the edges with valid occupancy update the path and push it away from the obstacles. The images at $n=50$ show the planner status a few iterations before convergence. The path clears all obstacles, however process has not yet converged since opposing objective functions, motion and obstacles, has not equalised yet. After 59 iterations the algorithm has converged to its final solution. With a mini-batch of 20 samples per iteration about 1200 samples were used in order to reach convergence. Deterministic sampling methods such as \cite{Marinho2016,mukadam2016gaussian,Zucker2013} require dense sampling, in the order of 100s of samples per iteration, of the objective function to decide on the best update location. Hence the stochastic path planner offers significant reduction in computational costs.

\subsection{Real data}\label{subsec:real_data}

\begin{figure*}[bt]
	\centering
 	
    \begin{subfigure}{0.3\textwidth}
        \includegraphics[width=\textwidth]{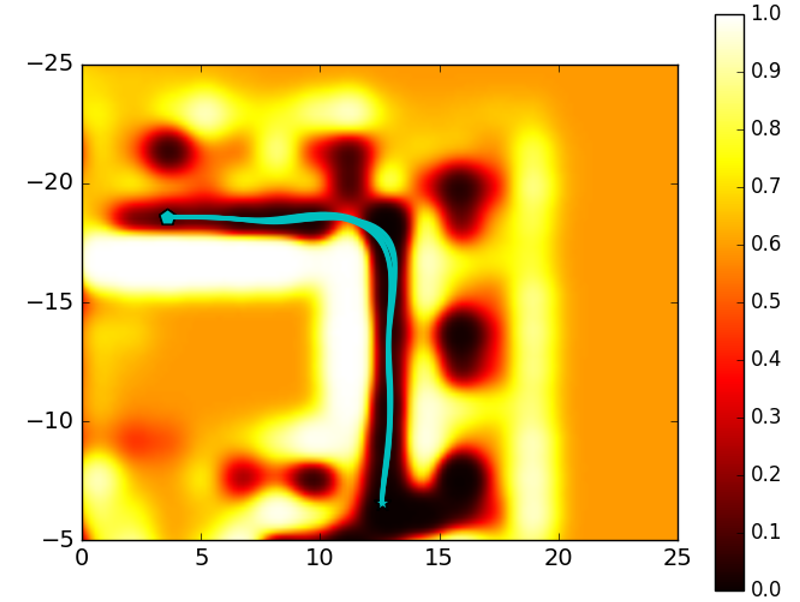}
        \caption{Approximate Kernel}
        \label{Intel_Comparison:subfig-1}
    \end{subfigure}
     \begin{subfigure}{0.3\textwidth}
        \includegraphics[width=\textwidth]{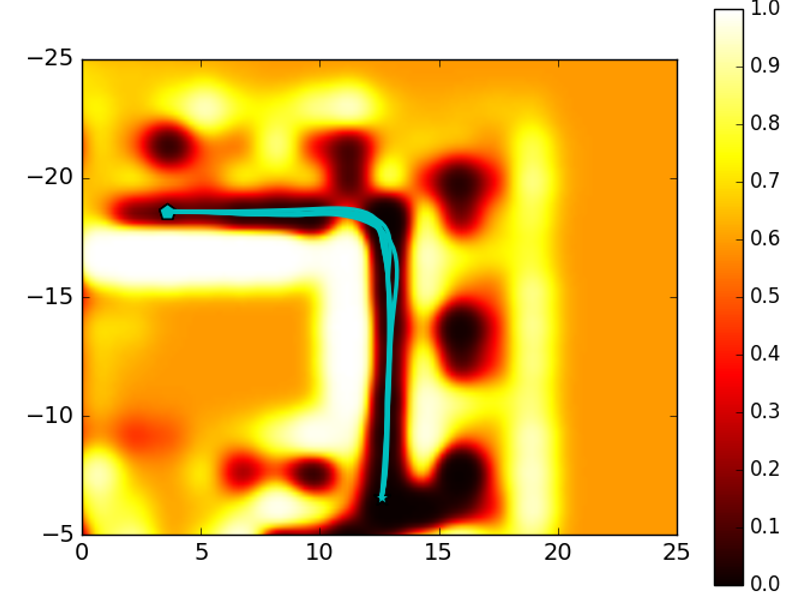}
        \caption{GP Paths \cite{Francis2017}}
        \label{Intel_Comparison:subfig-2}
    \end{subfigure}
    \begin{subfigure}{0.3\textwidth}
        \includegraphics[width=\textwidth]{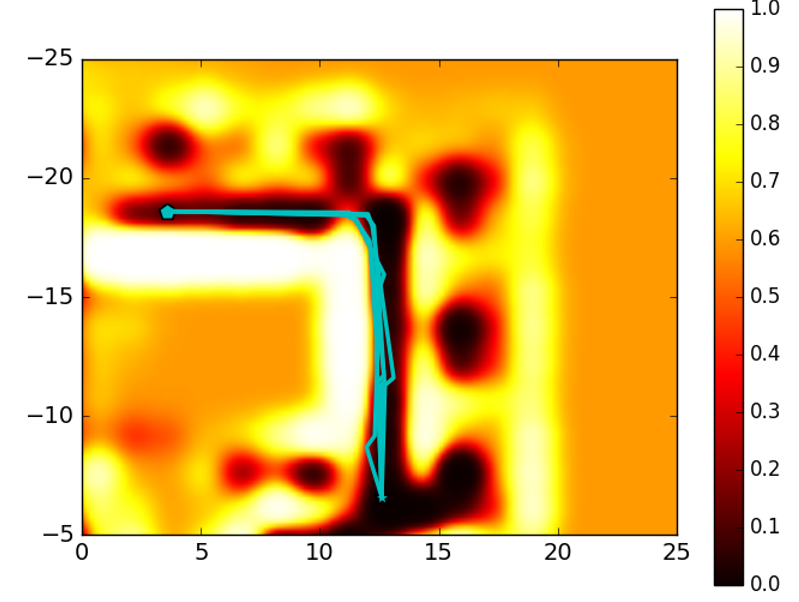}
        \caption{$\text{RRT}^*$ \cite{karaman2010incremental}}
        \label{Intel_Comparison:subfig-3}
    \end{subfigure}
\par\medskip
    
 	\caption{Comparison of path planning methods on a continuous occupancy map of the Intel-Lab; \protect\subref{Intel_Comparison:subfig-1} Our method, \protect\subref{Intel_Comparison:subfig-2} stochastic non-parametric GP paths \cite{Francis2017} and \protect\subref{Intel_Comparison:subfig-3} $\text{RRT}^*$ \cite{karaman2010incremental}. Each image shows five paths generated by the planning algorithm, to indicate repeated performance. Both stochastic methods \protect\subref{Intel_Comparison:subfig-1} and \protect\subref{Intel_Comparison:subfig-2} produce smooth paths which follow the mid lines between walls. $\text{RRT}^*$ paths are typically not smooth, and some have small jerks. In addition, $\text{RRT}^*$ paths are dangerously close to the walls.}
 	\label{fig:Intel_Comparison}
 \end{figure*}

The map for this experiment is based on the Intel-Lab dataset (available at http://radish.sourceforge.net/). We compare the optimal trajectory of our proposed method with two other methods; $\text{RRT}^*$ \cite{karaman2010incremental} and the stochastic GP path planner \cite{Francis2017}. Fig. \ref{fig:Intel_Comparison} and Table \ref{tab:performance_comparison} show a comparison between the different methods. $\text{RRT}^*$ forms a path based on several waypoints (states) the robot should pass from start to goal. As a result, the path typically is jagged, with short jerks. In contrast, the path generated by our method is continuous and smooth. In addition, unless using inflated obstacles, $\text{RRT}^*$ paths tend to move close to the walls or undershooting corners, as indicated by the relative high, and unsafe, occupancy of $\text{RRT}^*$ in Table \ref{tab:performance_comparison}. The stochastic planner follows the mid line between obstacles and perform smooth turns resulting in shorter and safer trajectories \footnote{Visualisation of the optimisation process is available at https://youtu.be/uf0qFWFJ83k}. 

\begin{figure}[tb]
 \centering
 	 \includegraphics[width=0.35\textwidth]{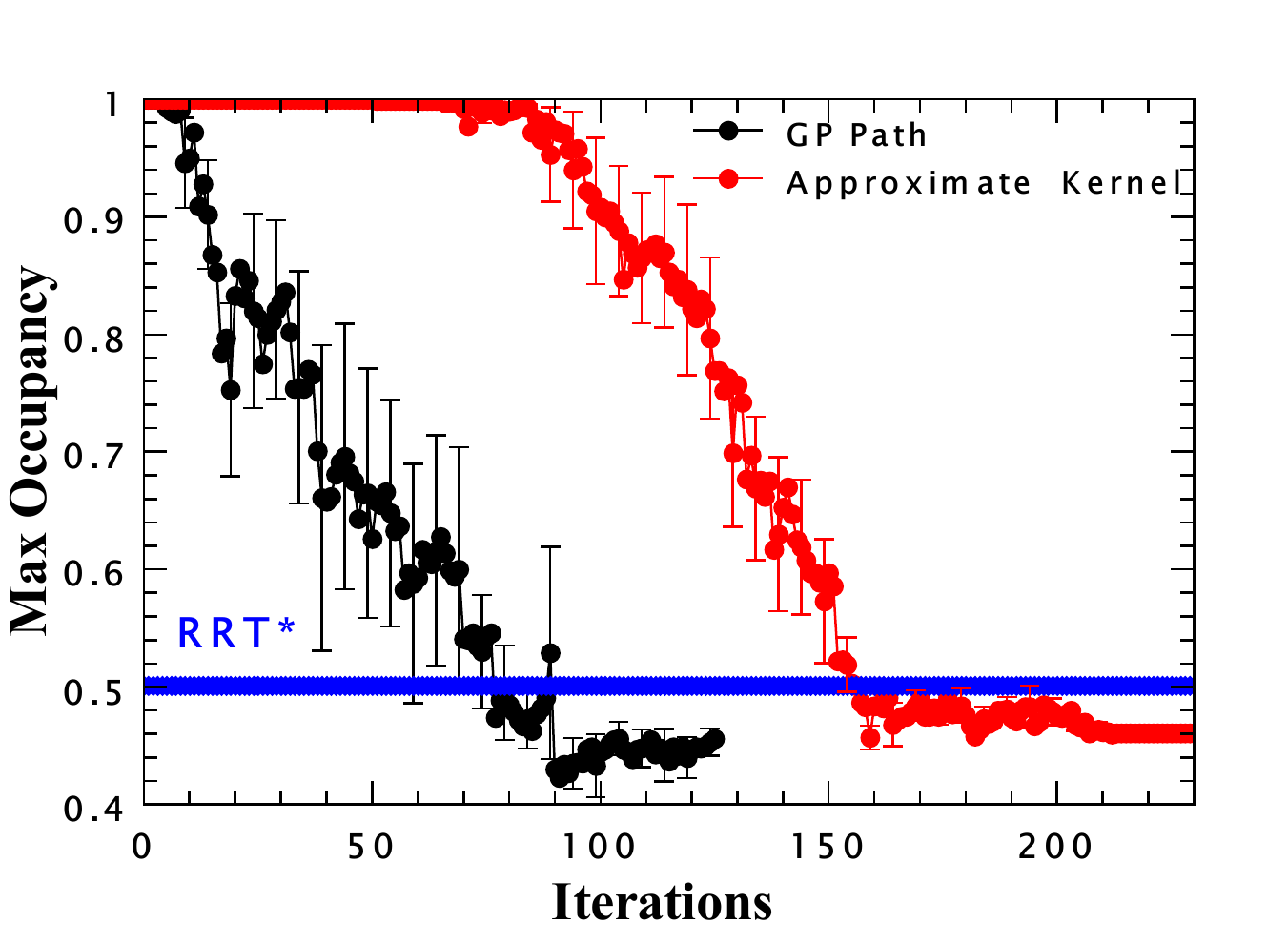}
 	\caption{Convergence comparison of the two stochastic functional gradient motion planners shown in Fig. \ref{fig:Intel_Comparison}. The maximum occupancy along the trajectory is plotted as a function of the iteration. The average and standard error over 5 repetitions is shown. The blue dashed line marks the average performance of $\text{RRT}^*$. Our proposed method requires more iterations than the GP method. However, in practice the runtime of our proposed method is lower, as the cost of adding new observations is fixed and linear in the number of features, while the GP path planner's computational cost is cubic in the number of samples.}
 	\label{fig:Intel_convergence}
\end{figure}

Qualitatively, both stochastic planners (\ref{Intel_Comparison:subfig-1} and \ref{Intel_Comparison:subfig-2}) present similar paths, as indicated in Table \ref{tab:performance_comparison} by the similar maximum occupancy and length. However, quantitative comparison of convergence between the methods shown in Fig. \ref{fig:Intel_convergence}, reveals a difference. Fig. \ref{fig:Intel_convergence} depicts the maximum occupancy along the trajectory as a function of iteration. The occupancy drops consistently as the update process progresses, until an occupancy of 0.45 is reached which corresponds to the occupancy in the vicinity of the corner. Both stochastic planners exhibit consistent performance in repeated experiments, even though they employ stochastic updates. However, the stochastic non-parametric GP planner of \cite{Francis2017} requires less iteration to converge. According to Table \ref{tab:performance_comparison}, the stochastic GP path planner requires, on average, a third of the samples used by the approximate kernel planner. This is mainly due to the highly responsive path model formed by the non-parametric GP path representation. Yet, using GPs for path representation limits scalability of this planning method. The main impediment of the GP planner is its cubic computational complexity. With more observations, updating and querying the GP path model becomes the bottleneck of the optimisation process. In contrast, our method uses an approximate kernel path representation, which has a fixed linear complexity. Consequently, adding more observations does not change the computational performance of the model. As a result, the time need to obtain a solution of the proposed method is much shorter compared with the GP planner. 

~

\begin{table}[thbp]
	\centering
	\caption{Performance comparison}
	\label{tab:performance_comparison}
	\begin{tabular}{lccc}
		\toprule
		& Approx. Kernel &
        Gp Paths \cite{Francis2017} &
        $\text{RRT}^*$ \cite{karaman2010incremental}  \\
		\midrule
		Max. occupancy & $ 0.46 \pm 0.01$ & $ 0.45 \pm 0.02$ & $ 0.50 \pm 0.01$ \\
		Path length [m]   & $20.48 \pm 0.07$ & $20.55 \pm 0.06$ & $20.17 \pm 0.14$ \\
        Samples   & $3530 \pm 160$ & $1080 \pm 250$ & $12180 \pm 650$ \\
        
        \bottomrule
	\end{tabular}
\end{table}

\section{Conclusions}\label{sec:conclusions}

The planning method proposed in this work employs SGD to optimise a path represented by an approximate kernel feature set. This model provides a highly expressive path in a cost effective representation. SGD combines the approximate kernel path model with a stochastic sampling schedule to form a computationally efficient optimisation process with convergence guarantees.

Planning in occupancy maps is a challenge for trajectory optimisers. Occupancy maps are a product of sensor observations and thus have contextual information gaps in the map due to lack of observations or occlusions. As a result, the path can not be optimised around these areas. Using stochastic samples across the entire path domain avoids the need to commit to an a-priori sampling resolution of the objective function. Consequently, the optimiser identifies transition areas around the obstacles borders, which enables the optimiser to overcome the gaps formed by the obstacles. 

Experimental results, in simulation and with real data, demonstrates the importance of stochastic sampling for planning in occupancy maps. Combined with an approximate kernel path representation, our method offers a scalable and fast method for trajectory optimisation in occupancy maps.



\bibliographystyle{ieeetr}
\bibliography{ApproxKernelPathPlanner}
\end{document}